\documentclass[10pt,letterpaper]{article}

\usepackage{amsmath} 

\renewenvironment{abstract}
{
  \centerline
  {\large \bfseries \scshape Abstract}
  \begin{quote}
}
{
  \end{quote}
}

\usepackage[top=1in,left=1in]{geometry}

\usepackage[utf8]{inputenc}

\usepackage{cite}

\usepackage{nameref,hyperref}

\usepackage{microtype}
\DisableLigatures[f]{encoding = *, family = * }

\raggedright
\usepackage{changepage}

\usepackage[aboveskip=1pt,labelfont=bf,labelsep=period,singlelinecheck=off]{caption}

\makeatletter
\renewcommand{\@biblabel}[1]{\quad#1.}
\makeatother

\usepackage{color}

\definecolor{Gray}{gray}{.25}

\usepackage{graphicx}

\usepackage{sidecap}

\usepackage{wrapfig}
\usepackage[pscoord]{eso-pic}
\usepackage[fulladjust]{marginnote}
\reversemarginpar

\begin{document}
\vspace*{0.35in}

\begin{flushleft}
{\Large
\textbf\newline{\LaTeX \textbf{AI Adoption to Combat Financial Crime: Study on Natural Language Processing in Adverse Media Screening of Financial Services in English and Bangla multilingual interpretation}
}
\newline
\\
\centerline{\small{\bfseries Soumita Roy}\textsuperscript{1,*}
}
\centerline{\textsuperscript{1}\small{Institute of Business Administration - JU}}
\centerline{\textsuperscript{*}\small20203552355roy@juniv.edu
}}
\end{flushleft}
\bigskip

\begin{abstract}
This document explores the potential of employing Artificial Intelligence (AI), specifically Natural Language Processing (NLP), to strengthen the detection and prevention of financial crimes within the Mobile Financial Services(MFS) of Bangladesh with multilingual scenario. The analysis focuses on the utilization of NLP for adverse media screening, a vital aspect of compliance with anti-money laundering (AML) and combating financial terrorism (CFT) regulations. Additionally, it investigates the overall reception and obstacles related to the integration of AI in Bangladeshi banks. This report measures the effectiveness of NLP is promising with an accuracy around 94\%. NLP algorithms display substantial promise in accurately identifying adverse media content linked to financial crimes. The lack of progress in this aspect is visible in Bangladesh, whereas globally the technology is already being used to increase effectiveness and efficiency.  Hence, it is clear there is an issue with the acceptance of AI in Bangladesh. Some AML \& CFT concerns are already being addressed by AI technology. For example, Image Recognition OCR technology are being used in KYC procedures. Primary hindrances to AI integration involve a lack of technical expertise, high expenses, and uncertainties surrounding regulations. This investigation underscores the potential of AI-driven NLP solutions in fortifying efforts to prevent financial crimes in Bangladesh. 

\end{abstract}

\section*{1. Introduction}
The report is based on the work experience acquired at an MFS. Adverse Media Screening is one of the scopes where AI can have a significant impact on preventing financial crimes. This process involves identifying negative information related to individuals or entities, such as news articles, court records, or social media posts, that could indicate potential involvement in illicit activities. Conventional methods of adverse media screening are often time-consuming and labor-intensive. AI, particularly NLP, can automate this process, making it more efficient and effective. One key analysis as part of NLP, Sentiment analysis, particularly is being researched for identifying negative information related to anti money laundering concerns (Han et al. 2020).

The various technologies are being used for AML and CFT (Combating Financial Terrorism) concerns as noted by researchers from the armed forces of other countries (T.H. Helmy et al., 2014). So, it is clear technology, finance and compliance concerns can come together to create solutions for this issue.

Sentiment analysis particularly has shown an accuracy rate of $76.96 \%$ for the research on 12,467 financial articles classifying (Han et al. 2020). That means accuracy of the contents are likely to be red flagged and predicted by the AI model for financially risky transactions.

A comprehensive research on the NLP uses of sentiment analysis notes use cases and researches for Product Analysis, Market Research, Competitor Analysis, Reputation Management, Business Intelligence, Health Care, Review Analysis, Experience Analysis, Sentiments of Customer Reviews, Aspect Analysis, Stock Market, Stock Price Prediction, and Trend Prediction (Wankhade et al., 2022). While the analysis at the document level focuses on classifying the entire text as either positively or negatively subjective,\\
examining individual sentences proves to be more useful since a document can include both positive and negative remarks (Wankhade et al., 2022).

Another research on 20000 sample records from publicly available datasets found using Google's AutoML tool found 85 percent accuracy (M. H. Thi et al., 2020). The research also notes the usability of such technology to allow information availability of new clients. This level of accuracy suggests that the potential of using sentiment analysis from commercially available tools (M. H. Thi et al., 2020).

The use of classification confusion matrix modeling for predicting AML transactions' risk accuracy, precision, recall and f 1 score has been researched (M. Alkhalili et al. 2021). The Research also notes the riskiness of False Positive and False Negative, as False Positive is more risky since letting a financially negative news go without investigation will put the financial institutions at risk (M. Alkhalili et al. 2021). The research further suggests the higher the accuracy the better. An accuracy of $50 \%$, which is likely to be the least acceptable accuracy benchmark, is no better than a coin toss of random guesses.

Additionally 90 percent accuracy was found for classification of users' sentiments of extremist terrorist concern which is a concern of AML and combating financial terrorism (Ahmad, S. F.M. et al., 2019).

IBM research notes a 30 percent reduction in investigation time, which allows the financial investigator efficiency to investigate more cases while reducing administrative and human resource costs associated with financial investigation (Weber, M. et al., 2018).

\section*{2. Method}
This research has used methods for the NLP effectiveness in sentiment measures of the positive, negative and neutral connotation of the data. Tools used include PyCharm, Open AI Large language Model, and Python library. Raw data is collected from data sources, which is dependent upon data sources.

The use of classification confusion matrix method for predicting AML transactions' risk accuracy, precision, recall and f1 score has been researched for the case of AML (M. Alkhalili et al. 2021). Below is the formula for quantifying accuracy, precision, recall and F1 score.

\begin{verbatim}
Accuracy = (TP+TN) /(TP+FP+TN+FN)
Precision = TP /(TP+FP)
Recall = TP /(TP+FN)
F1-score = 2 * (Precision * Recall) / (Precision + Recall)
\end{verbatim}
 Given that,\\
TP = True Positive (Correctly identified positive instances)\\
TN= True Negative(Correctly identified negative instances)\\
FP = False Positive(Incorrectly identified positive instances)\\
\textit{FN= False Negative(Incorrectly identified negative instances)}\\

TP, FP, TN, and FN are used to calculate Precision, Recall, F1 Score and Accuracy:

\begin{itemize}
  \item Precision emphasizes how many of the positive predictions were accurate (pertaining to TP and FP).
  \item Recall assesses how effectively the model identifies all genuine positive cases (pertaining to TP and FN).
  \item F1 Score merges precision and recall to provide a balanced evaluation.
  \item Accuracy is a very simple and widely used measure to benchmark many AI models and tools. It is all the accurate prediction outcomes divided by all the data.
  \item Weighted measures of this calculation are based on the data sets weights of positive, neutral and negative data.
\end{itemize}

Accuracy, Precision, recall, and the F1 score are metrics utilized as an evaluation measure of performance of classification models, particularly in situations where class imbalance exists or when there's an emphasis on particular types of errors.

\subsection*{2.2 SOURCES OF DATA AND SAMPLING}
Research population data sources are in social media and news, as this is where adverse media or negative media can be found. This research samples the data from mainly facebook, as it is widely used for money laundering and financial crimes, and news articles are not from specific news companies rather from all the digital news articles published during the sampling time discussed later in the report. Sample size is 5376 text sentences and fragments from the sources mentioned. Data was collected over the first week of September. So, the time frame includes a week's news from the media related to MFS.

\section*{3. Experiment}
Data collection was automatedly collected by using Data scraping. Data cleaning was also taken into account, which was to prepare the data into text formatting and removing unnecessary metadata, images and their title. Afterwards some of the data had to be manually labeled if they are positive, neutral or negative. Since the data was labeled data we could evaluate the models in supervised learning to utilize classification analysis evaluating the accuracy, precision, and recall.

\textbf{Training data splitting}:\\
Research also suggests $2 / 3$ data splitting ratio of training to testing had better outcomes than $1 / 3$ data splitting (Dobbin, K. K., \& Simon, R. M., 2011). This testing has roughly the following data splitting.

\begin{itemize}
  \item Training Documents Percentage : $64.22 \%$
  \item Testing Documents Percentage : $35.78 \%$
\end{itemize}

Table 1: Coded Data Classes of the Data Collection

\begin{center}
\begin{tabular}{|l|l|}
\hline
Positive & 133 \\
\hline
Neutral & 5022 \\
\hline
Negative & 221 \\
\hline
Total & $\mathbf{5 3 7 6}$ \\
\hline
\end{tabular}
\end{center}

The limited training on the 5376 text sentences/fragments from various sources including Social Media, facebook, and news media of a week. Central tendency is observed in the data. Data is kept in sentence/fragment, as examining individual sentences proves to be more useful since a document can include both positive and negative remarks (Wankhade et al., 2022).

\section*{4. Result}
\subsection*{4.1 FINDING AND ANALYSIS}
In the data samples below, it shows some of the samples of the analysis. True positive is interchangeably for the all the sentiments predicted correctly, as the $3 * 3$ confusion matrices have positive, negative and neutral classification, for which true negative is in other words the true positive cases of negative sentiments.

Table 2: Confusion Matrix of the classification of the testing dataset

\begin{center}
\begin{tabular}{cc|c|c|c}
 &  & \multicolumn{3}{c}{Labeled as (Actual Data)} \\
\hline
 & & negative& neutral &  positive \\
 & negative& 50 & 23 &2 \\
\hline
\begin{tabular}{r}
Predicted as \\
(AI Data) \\
\end{tabular} & neutral & 32& 1603 & 16 \\
\hline
 & positive & 0 & 23& 17\\
\hline
\end{tabular}
\end{center}

\begin{center}
\includegraphics[scale=.3]{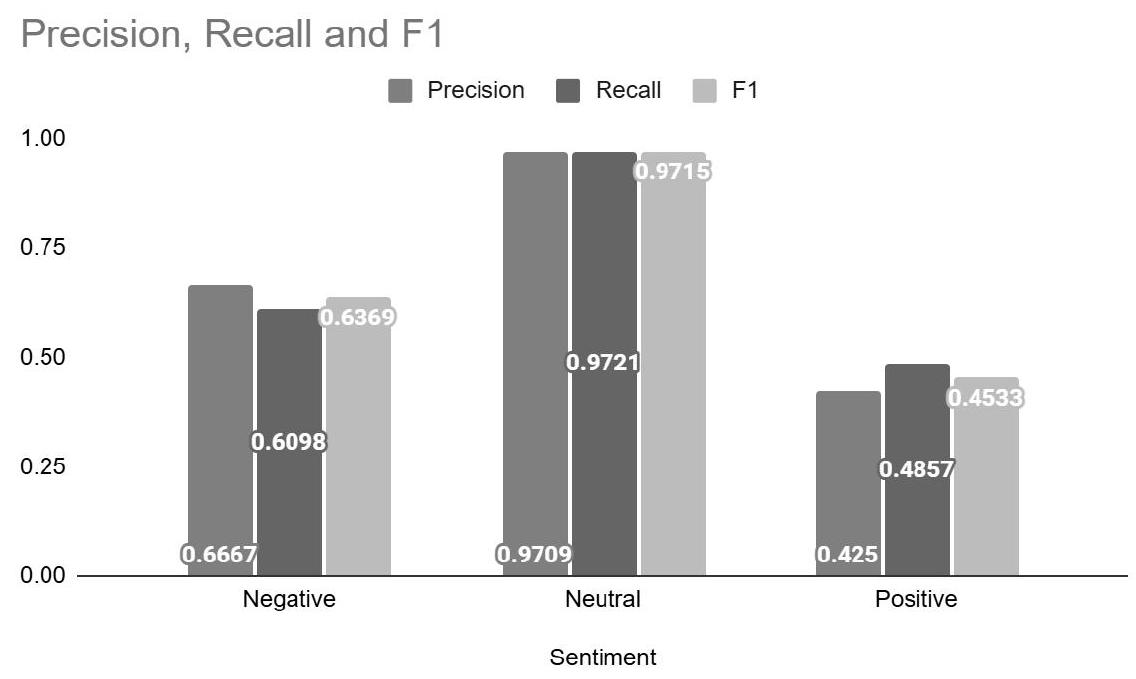}
\end{center}

Figure 1: Precision, Recall and F1 score for each sentiments (negative, neutral, positive)\\
The most valuable precision for the use case of AML and CFT concern is the negative precision. The model correctly predicts negative cases 66 percent of the time, which is significant as the model will need to find negative articles and news for AML concerns. Capability of the negative outcomes are more concerning then positive outcomes as suggested by some research (M. Alkhalili et al. 2021). F1 shows the harmonic balance of each class, so it is like an average measure of recall and precision.

For positive sentiment recall and precision are showing poor predictability, less than 50 Percent. This can further be improved with more training data, but this positive case is not as necessary for the AML use case.

\subsection*{4.2. WEIGHTED PRECISION, RECALL, F1 SCORE}
To determine the weighted precision, we must consider the precision of each class along with the number of instances (support) related to that class. The formula to compute weighted measures is:

Weighted Measure:
\[
\text{Weighted Measure} = \frac{\Sigma \text{ Measure for each class } \times \text{ Support for that class }}{\Sigma \text{ Support for all classes }}
\]

Table 3: Confusion Matrix of the classification of the testing dataset

\begin{center}
\begin{tabular}{|l|l|l|l|}
\hline
Accuracy & Weighted Precision & Weighted Recall & Weighted F1 Score \\
\hline
0.9456 & 0.9460 & 0.9458 & 0.9456 \\
\hline
\end{tabular}
\end{center}

All of the weighted scores are around $94 \%$, which means the weighted average shows the model is feasible. However, the weight for neutral data is the highest, which impacts the weight of the score. Hence, we should instead look at the precise score of each sentiment discussed earlier.

\subsection*{4.3. ACCURACY}
Accuracy is a widely used metrics for measuring efficacy of AI models. The formula for Accuracy is:\\
Accuracy $=\frac{\text { All Correct Prediction }}{\text { All Prediction }}$

In other words,

Accuracy $=\frac{\text { True Positives }+ \text { True Negative }}{\text { True Positives }+ \text { False Positives }+ \text { True Negative }+ \text { False Negative }}$

\textbf{Accuracy: 0.9456}
The model correctly predicts $\mathbf{9 4 . 5 6 \%}$ of the times.\\
\begin{center}
\includegraphics[scale=.2]{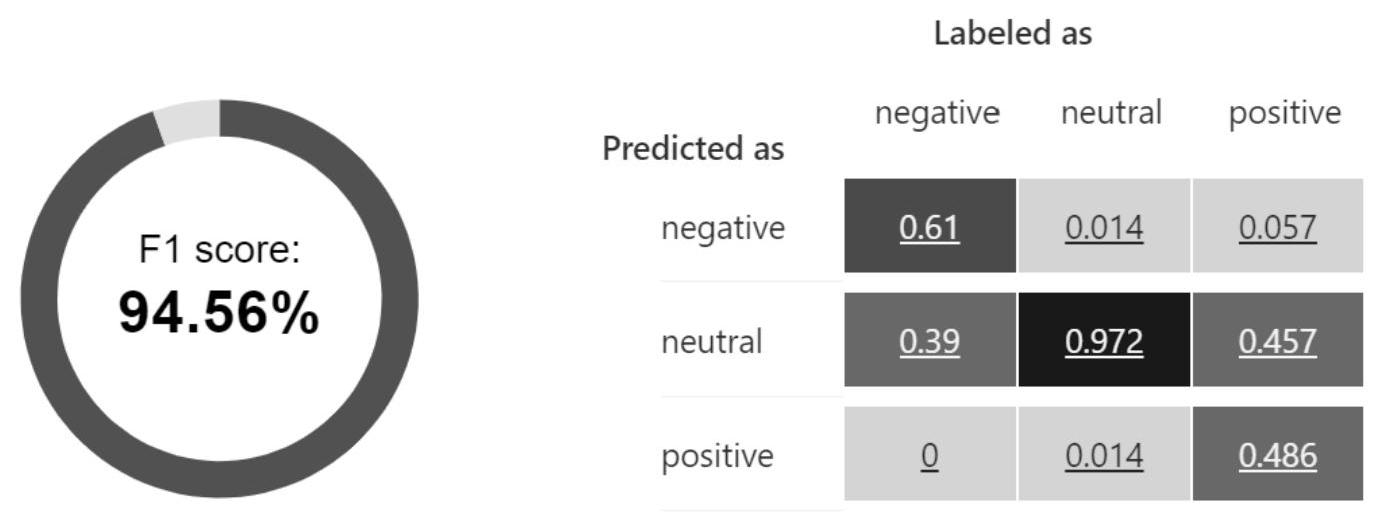}
\end{center}
Figure 2: F1 Score and Classification

Overall, the model is showing the NLP method to be feasible for predicting adverse media screening. The model correctly predicts $94.56 \%$ of the times, according to the accuracy score. Additionally, the F1 score of $94.56 \%$ shows the mode is especially useful when there is a need to maintain a balance between precision and recall. Neutral data prediction is most accurate of the three, as the data available for that was also higher. Negative data is predicted with more accuracy. Positive data is predicted with less accuracy. However, data are very less likely to be labeled positive when in reality it is negative and vice versa with the likelihood of $5.7 \%$ and $0 \%$ according to the classification ratio (Figure 2).

Obstacles faced by AI models like this are many, but the opportunity it presents is also huge. Firstly, AI models alone are not enough, human intervention is needed in every step of the way. AI could also be impacted by human judgement and bias, so error free models are very difficult to mandate. This model also shows that even though the negative data can be found with the model, which one deserves further action can be judged by an expert in the field based on the nuance of laws and policy of the country. Secondly, this model alone is not enough. There is a need to add other methods to make the model even more effective, however, there is always a scope for improvement.

\section*{CONCLUSION}
To simply put, there is a lot of potential for AI in the financial sector, but the adoption is very challenging for multitudes of reasons. Without proper risk mitigation of using AI, it is not possible to implement it without human intervention.

Compliance needs to be more efficient and effective to safeguard both customers and the financial institutions. AI adoption has a proven track record of working for the benefit of the customer, and in the coming years it will be required to strengthen more on it. The study is specifically confined to the MFS industry of Bangladesh and does not encompass other sectors or geographical areas.

\section*{REFERENCES}

Han, J., Huang, Y., Liu, S. et al. (2020) Artificial intelligence for anti-money laundering: a review and extension. Digit Finance 2, 211-239. \href{https://doi.org/10.1007/s42521-020-00023-1}{https://doi.org/10.1007/s42521-020-00023-1}\\

Goutte, Cyril \& Gaussier, Eric. (2005). A Probabilistic Interpretation of Precision, Recall and. 345-359.\\
T.H. Helmy, M. Zaki, T. Salah, and B.K. Tarek. 2014. Design of a monitor for detecting money laundering and terrorist financing. Journal of Theoretical and Applied Information Technology, 1(1):1-11.\\

Dobbin, K. K., \& Simon, R. M. (2011). Optimally splitting cases for training and testing high dimensional classifiers. BMC medical genomics, 4, 31. \href{https://doi.org/10.1186/1755-8794-4-31}{https://doi.org/10.1186/1755-8794-4-31}\\

M. H. Thi, C. Withana, N. T. H. Quynh and N. T. Q. Vinh, (2020) "A Novel Solution For Anti-Money Laundering System," 2020 5th International Conference on Innovative Technologies in Intelligent Systems and Industrial Applications (CITISIA), Sydney, Australia, pp. 1-6, doi: 10.1109/CITISIA50690.2020.9371840.\\

M. Alkhalili, M. H. Qutqut and F. Almasalha, (2021) "Investigation of Applying Machine Learning for Watch-List Filtering in Anti-Money Laundering," in IEEE Access, vol. 9, pp. 18481-18496, doi: 10.1109/ACCESS.2021.3052313.\\

Ahmad, S., Asghar, M.Z., Alotaibi, F.M. et al. (2019) Detection and classification of social media-based extremist affiliations using sentiment analysis techniques. Hum. Cent. Comput. Inf. Sci. 9, 24. \href{https://doi.org/10.1186/s13673-019-0185-6}{https://doi.org/10.1186/s13673-019-0185-6}\\

Weber, M., Chen, J., Suzumura, T., Pareja, A., Ma, T., Kanezashi, H., ... \& Schardl, T. B. (2018). Scalable graph learning for anti-money laundering: A first look. arXiv preprint arXiv:1812.00076, 1-7.\\

Wankhade, M., Rao, A.C.S. \& Kulkarni, C. (2022) A survey on sentiment analysis methods, applications, and challenges. Artif Intell Rev 55, 5731-5780. \href{https://doi.org/10.1007/s10462-022-10144-1}{https://doi.org/10.1007/s10462-022-10144-1}\\

\end{document}